\documentclass[lettersize,journal]{IEEEtran}
\usepackage{amsmath,amsfonts}
\usepackage{array}
\usepackage[caption=false,font=normalsize,labelfont=sf,textfont=sf]{subfig}
\usepackage{textcomp}
\usepackage{stfloats}
\usepackage{url}
\usepackage{verbatim}
\usepackage{graphicx}
\def\BibTeX{{\rm B\kern-.05em{\sc i\kern-.025em b}\kern-.08em
		T\kern-.1667em\lower.7ex\hbox{E}\kern-.125emX}}
\usepackage{balance}

\usepackage{graphicx}
\usepackage{amsmath}
\usepackage{amsthm}
\usepackage{booktabs}
\usepackage{algorithm}
\usepackage{algorithmicx}
\usepackage{algpseudocode}
\usepackage[switch]{lineno}

\usepackage{float}
\usepackage{subfig}
\usepackage{amssymb}
\usepackage{bm}
\usepackage{indentfirst}
\usepackage{color}
\usepackage{epsfig}
\usepackage{mathtools}

\usepackage{lineno}
\usepackage{bbm}
\usepackage[capitalize]{cleveref}

\usepackage{multirow}
\usepackage[top=2cm, bottom=2cm, left=2cm, right=2cm]{geometry}

\usepackage{diagbox}

\begin{document}
	\title{
		Multi-Level Label Correction by Distilling Proximate Patterns for Semi-supervised Semantic Segmentation
	}

	\author{Hui Xiao, Yuting Hong, Li Dong, Diqun Yan, Jiayan Zhuang, Junjie Xiong, Dongtai Liang and $\text{Chengbin Peng}^{*}$
		\thanks{ $*$ Corresponding author.
			
			Hui Xiao, Yuting Hong, Li Dong, Diqun Yan, Dongtai Liang and Chengbin Peng are with the Faculty of Electrical Engineering and Computer Science, Ningbo University, Ningbo 315211, China (email: 2011082337@nbu.edu.cn; 2211100255@nbu.edu.cn; dongli@nbu.edu.cn; yandiqun@nbu.edu.cn; liangdongtai@nbu.edu.cn; pengchengbin@nbu.edu.
			cn).(Corresponding author:Chengbin Peng.)
			
			Jiayan Zhuang is with Ningbo Institute of Materials Technology and Engineering, Chinese Academy of Sciences, Ningbo 315201, China (email: zhuangjiayan@nimte.ac.cn).
			
			Junjie Xiong is with Hangzhou Shenhao Technology Co., Ltd, Hangzhou 310000, China (email: Xiongjunjie@shenhaoinfo.com). 
			
	}}
	
	
	\maketitle
	
	\begin{abstract}
		Semi-supervised semantic segmentation relieves the reliance on large-scale labeled data by leveraging unlabeled data. Recent semi-supervised semantic segmentation approaches mainly resort to pseudo-labeling methods to exploit unlabeled data. However, unreliable pseudo-labeling can undermine the  semi-supervision processes. In this paper, we propose an algorithm called Multi-Level Label Correction (MLLC), which aims to use graph neural networks to capture structural relationships in Semantic-Level Graphs (SLGs) and Class-Level Graphs (CLGs) to rectify erroneous pseudo-labels.  Specifically,  SLGs represent semantic affinities between pairs of pixel features, and CLGs describe classification consistencies between pairs of pixel labels. With the support of proximate pattern information from graphs, MLLC can rectify incorrectly predicted pseudo-labels and can facilitate discriminative feature representations. We design an end-to-end network to train and perform this effective label corrections mechanism. 
		Experiments demonstrate that MLLC can significantly improve supervised baselines and outperforms  state-of-the-art approaches in different scenarios on Cityscapes and PASCAL VOC 2012 datasets.  
		Specifically, MLLC improves the supervised baseline by at least 5\% and 2\% with DeepLabV2 and DeepLabV3+ respectively under different partition protocols.
	\end{abstract}

	\begin{IEEEkeywords}
		Semantic segmentation, Semi-supervised learning, Pseudo label, Graph convolution.
	\end{IEEEkeywords}

	
	\section{Introduction}
	\IEEEPARstart{S}{emantic} segmentation aims to generate pixel-wise category predictions for an given image \cite{long2015fully,chen2017deeplab,kirillov2023segany} and can be applied to fields such as autonomous driving and medical image analysis. However, its performance heavily depends on the quantity and quality of annotated datasets, and the annotation process can be laborious and time-consuming. For example, annotating a single image of the Cityscapes dataset takes on average 1.5 hours \cite{cordts2016cityscapes}. Therefore, how to effectively use unlabeled data is an issue that is currently attracting widespread attention.
	
	\begin{figure}[!t]
		\begin{center}
			
			\subfloat{\includegraphics[width=1\linewidth]{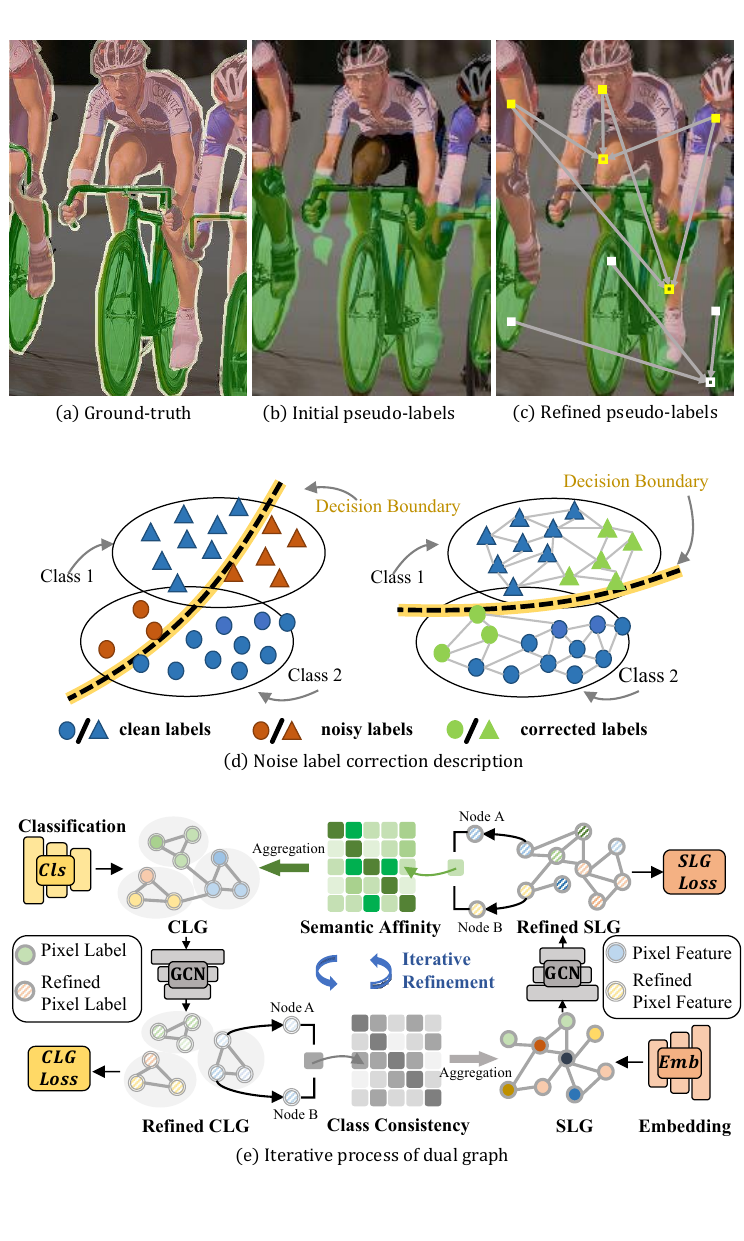}} \\

		\end{center}
		\vspace{-25pt}
		
		\caption{\textbf{A brief illustration.} (a) Ground-truth labels for pixels represented by GREEN and PINK, as a mask on input image. (b) Pseudo-labels without refinement. (c) Pseudo-labels refined by MLLC framework. (d) The left and right subplots represent prediction distributions with and without MLLC, respectively, and by correcting noisy pseudo-labels, the framework can help classifiers to find better decision boundaries. (e) The updating process of MLLC, in which CLG aggregates semantic knowledge from SLG and SLG aggregates classification information from CLG during iterative refinements. 
		}
		\label{brief}
		\vspace{-12pt}
		
	\end{figure}

	Semi-supervised learning can exploit massive unlabeled data to facilitate learning of network under limited labeled data \cite{tarvainen2017mean,chen2021temporal,li2020snowball}. We expect the network trained with unlabeled data to achieve results comparable to fully-supervised setting. Consistency regularization is a common method in semi-supervised learning \cite{french2019semi,olsson2021classmix} . It encourages the network to produce similar predictions for different augmented versions of a single unlabeled image. Another semi-supervised learning method, self-training \cite{zhu2020improving,wei2021crest,teh2021gist,ke2020three,mittal2019semi,hung2018adversarial}, could use the trained network to generate pseudo-labels of unlabeled images and use them for supervised learning.

	However, due to the inevitable error-prone prediction of unlabeled data, the network is prone to over-fitting incorrect predictions, which can lead to confirmation bias \cite{arazo2020pseudo}. To alleviate this problem, existing methods use thresholds to filter the predictions \cite{hung2018adversarial,mittal2019semi,DBLP:conf/iccv/0001X0GH21,olsson2021classmix}, by retaining high confidence predictions and discarding those with low confidence. 
	which wastes potentially correct pseudo-labels indiscriminately. Some methods correct erroneous pseudo-labels by learning auxiliary networks,  for example, an error correction network (ECN) that learns from differences between network predictions and ground truth annotations 
	\cite{mendel2020semi,ke2020guided}. Yet, in practice, such networks can easily suffer from over-fitting when trained with a few labeled data only, and it is difficult to cover a wide variety of prediction errors.
	{
		In other learning paradigms, there are also a few  pseudo label correction approaches based on  graphs. In weakly supervised learning, Zhang et. al. \cite{zhang2021affinityattention}  use a single graph to propagate semantic labels  from confident pixels to unlabeled ones. In unsupervised learning, Yan et. al. \cite{yan2022plug} build a nearest neighbor  graph and propagate features on this graph to predict node linkages. Yet, these approaches generally rely on a single graph using very basic feature propagation methods, and they are not designed for semi-supervised learning. 
	}

	{To overcome these limitations, in this work, we propose a novel method that is dedicated to correct errors on pseudo-labels, as shown in Fig. \ref{brief}. Our motivation derives from the smoothness assumption \cite{luo2018smooth} that, for two samples close to each other in a dense data region, their class labels are similar. That is, samples from the same class should have higher similarity in the feature space than samples from different classes. 
	Thus, noisy labels can be corrected by aggregating information from nearby samples in the feature space according to the smoothness assumption, and similarly, relations among features belonging to the same class can be strengthened by globally aggregating features of samples with the same class \cite{song2022graph}, which can produce  more representative features.

	The core part of our approach is a multi-level label correction network, which aims to capture structural relationships between class space and feature space. Specifically, it consists of two parts, a semantic-level graph (SLG) representing semantic affinity between features and a class-level graph (CLG) representing consistency between predicted labels. To train the network, we design an end-to-end training paradigm that leads two graphs to interact and co-evolve. For potentially incorrect label predictions, the predictions can be improved by weighting it using corresponding semantic neighbors in the SLG. For feature representations, structure is regularized by aggregating neighbors of same category through CLG to obtain accurate representations. The training process is performed until converge, and we can obtain refined segmentation maps and smooth feature representations.
		}
	
	To construct reliable SLG and CLG, we propose two corresponding losses. The SLG loss is a contrastive loss containing both intra-image and cross-image comparisons and the SLG loss is a cross-entropy loss containing dynamic weighting coefficients. Specifically, The SLG loss allows samples with the same pseudo-labels to be similar after feature embedding, which is consistent with the smoothness assumption \cite{gong2015deformed}. The CLG loss can perform self-supervision for generated pseudo-labels, and assign smaller weights to potentially incorrect pseudo-labels, thus further alleviating the influence of noisy labels. Experiments demonstrate that the two losses jointly contribute to the performance gain for semi-supervised segmentation.
	
	Our contribution can be summarized as follows:
	\begin{itemize} 
		\item We propose a novel dual-graph framework to improve conventional semi-supervised learning. By exchanging structural information between feature space and label space, this approach can obtain refined segmentation maps and more representative features. 
		To the best of our knowledge, it is the first multi-level semi-supervised semantic segmentation approach based on graph convolutions. 
		\item A contrastive loss including both intra-image and cross-image comparisons is proposed to enhance inter-class feature discrepancy and intra-class feature compactness. In addition, a dynamic cross-entropy loss is proposed to further alleviate misleading of noisy pseudo-labels.
		
		\item We demonstrate state-of-the-art performance on multiple widely used benchmarks. The proposed method is evaluated on Cityscapes and PASCAL VOC 2012 datasets as convention, and experiments validate the superior performance on both datasets. 
	\end{itemize}

	\section{Related Work}
	\textbf{Supervised semantic segmentation} aims to assign labels for each pixels in an image. Since emergence of Fully Convolutional Networks (FCN) {\cite{long2015fully}}, various methods have been proposed which train a pixel-level classifier network. In pace with the development of semantic segmentation by designing different architectures, autonomous driving and medical imaging have made remarkable progress. Recent methods can be achieved excellent results by using encoder-decoder {\cite{chen2018encoder}}, ASPP {\cite{chen2017deeplab}} and PSPNet  {\cite{zhao2017pyramid}} structures. However, gathering the pixel-level labels is time-consuming and laborious, which limits their practical application. 
	
	\textbf{Semi-supervised semantic segmentation} aims to use both unlabeled and labeled data to train a high-performance segmentation network. Due to the limit of enough labels, how to make full use of unlabeled data becomes an important issue. Some preliminary works utilize Generative Adversarial Networks (GANs) {\cite{luc2016semantic}} as an auxiliary supervision signal for unlabeled data.  Recent works pay attention to consistency regularization and entropy minimization. FixMatch {\cite{sohn2020fixmatch}} combines two methods and obtained a satisfactory result. DMT {\cite{feng2022dmt}} applies two differently initialized models and uses disagreement between models to re-weight the loss function. C3-SemiSeg {\cite{zhou2021c3}} proposes a pixel-level contrast learning loss function to enhance the inter-class feature differences and intra-class feature compactness of a data set. By this way,  noisy pseudo-labels can lead to error accumulation and model degradation. How to use the pseudo-labels properly is a challenging task.
	Consistency-based semi-supervised learning utilizes the clustering assumption that decision boundaries typically pass through the region with low sample density  {\cite{ouali2020semi,xu2021dash,french2019semi}}. Different perturbations are added to input samples and models are expected to predict a consistent category. PSMT {\cite{liu2022perturbed}} proposes a data augmentation method including  perturbations in inputs, features, and networks to improve generalization of model. ClassMix {\cite{olsson2021classmix}} samples two images in the unlabeled data and makes mask copies for predictions and ground-truth respectively, which can introduce confirmation bias. 
	{  CCVC \cite{wang2023conflict} proposes a cross-view consistency approach enabling networks to learn useful information from unrelated views. Unimatch \cite{yang2023revisiting} proposes a dual-stream perturbation technique enabling two strong views and a weak view to produce consistent predictions.}
	Pseudo-labeling based semi-supervised learning considers predictions obtained from a reused model as supervision {\cite{lee2013pseudo}}, and the use of pseudo-labels is motivated by entropy minimization. Preliminary works usually set fixed thresholds to discard unreliable pseudo-labels. Based on FixMatch, FlexMatch  {\cite{zhang2021flexmatch}} introduces a curriculum pseudo-labeling that provides a dynamic threshold. AEL {\cite{hu2021semi}} maintains a dynamic memory bank to select high-quality pseudo-labels in different components. U2PL {\cite{wang2022semi}} fully uses unreliable pseudo-labels for contrastive learning to boost performance. 
	{ USRN \cite{guan2022unbiased} solves the class imbalance issue by learning class unbiased information. 
	GTA-Seg \cite{jin2022semi} uses an assistant network to disentangle effects of pseudo labels on feature extractor and mask predictor of student model.
	CISC-R \cite{wu2023querying} and GuidedMix-Net \cite{tu2022guidedmix} use knowledge from labeled data to guide learning from unlabeled data.
	CPCL \cite{fan2023conservative} uses consistency and inconsistency of the predictions of two networks for learning.
	FPL \cite{qiao2023fuzzy} uses contrastive learning to enable network is able to adaptively encourage fuzzy positive predictions and suppress highly probable negative predictions.
	DGCL \cite{wang2023hunting} and PCR \cite{xu2022semi} use contrastive learning to move features closer to center of clusters.
	AugSeg \cite{zhao2023augmentation} proposes a new data perturbation method to improve the performance of network.
}

	Rather than discarding low confidence pseudo-labels or using them selectively as many traditional approaches, in this work, we propose a dual-graph framework with carefully designed learning schemes, which can rectify incorrect noisy pseudo-labels more effectively.

	\section{Our Approach}
	In this section, we first define the problem and provide an overview of our approach. Then we introduce the core concept of the multi-level learning through dual graphs. 
	Finally, we describe how to train the dual graph.
	
	\begin{figure*}[t]
		
		\begin{center}
			
			\includegraphics[width=1.0\textwidth]{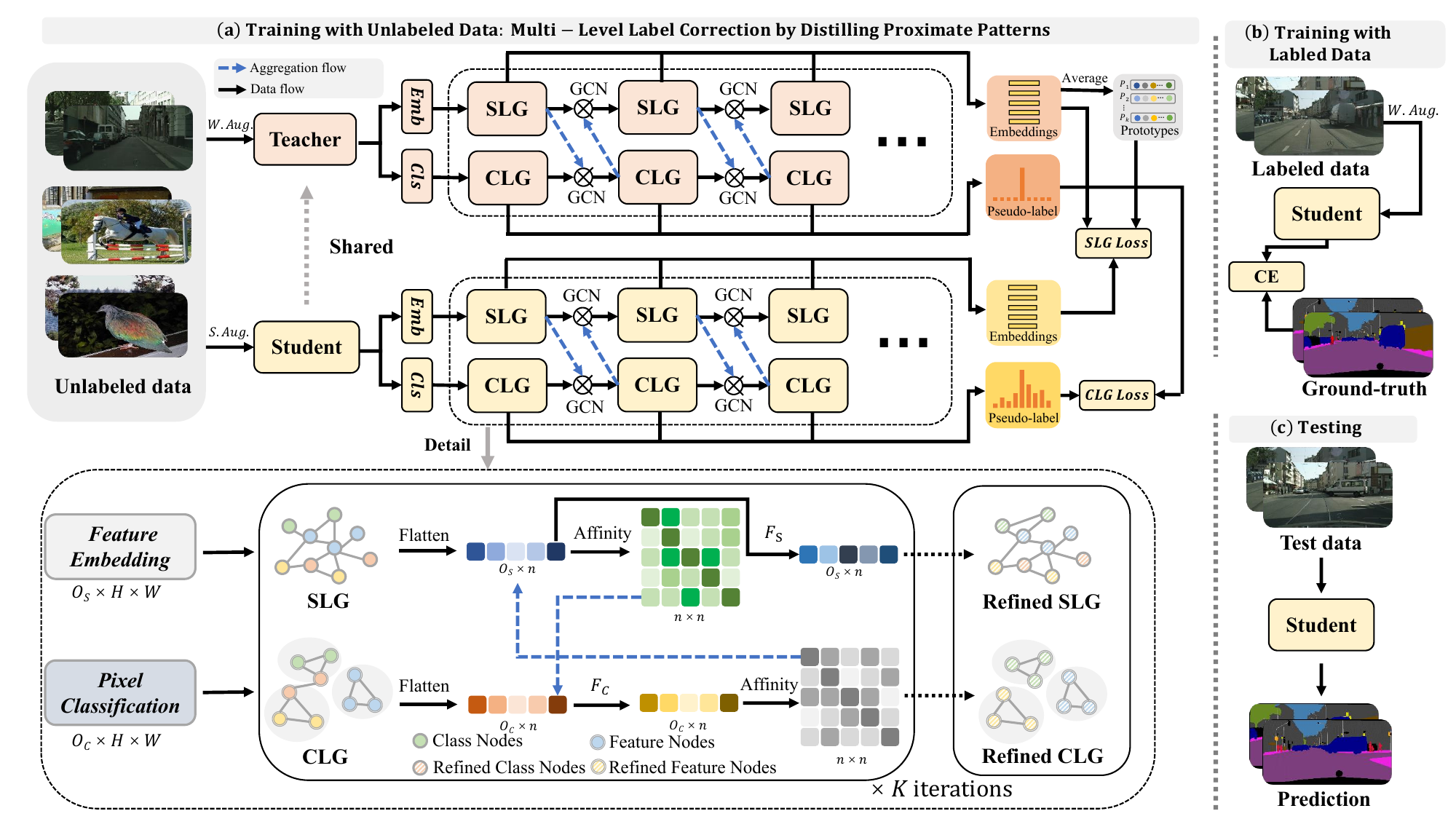}
			
		\end{center}
		
		\caption{\textbf{Overview of the proposed framework.} 	The network model consists of a CNN-based encoder and a decoder with two heads. One is a classification head represented by $Cls$ and a feature embedding head represented by $Emb$. For an unlabeled image, we first perform a strong data augmentation and a weak data augmentation represented by $S. Aug$. and $W. Aug.$ respectively, and then feed these two differently augmented data into student and teacher networks, respectively. Immediately following the network is a two-level graph framework designed to generate refined segmentation predictions, with $O_C$ and $O_S$ denoting the output size of the embedding head and the classification head respectively. 
			These more reliable predictions and features are then used as supervised knowledge for the student network. 
		}
		\label{fw}
		
	\end{figure*}
	
	\subsection{Preliminary}
	\textbf{Problem definition.}  Let $\mathcal{D}$ be a dataset that consists of a labeled set $\mathcal{D}_l=\{(X_{l}^{i},Y_{l}^{i})\}_{i=1}^{N_l}$ and a unlabeled set $\mathcal{D}_u=\{(X_{u}^{i}\}_{i=1}^{N_u}$, where $X^i\in{\mathbb{R}^{H\times{W}\times{3}}}$ is the $i$th input images with RGB channels and $Y^i\in{\mathbb{R}^{H\times{W}}}$ represents the corresponding pixel-level ground truth label. $N_l$ and $N_u$ denote the numbers of labeled samples and unlabeled samples respectively, and generally $N_l\ll{N_u}$. In our approach, labeled and unlabeled data are sampled equally in each training process. Semi-supervised semantic segmentation task aims to utilize both labeled $\mathcal{D}_l$ and unlabeled data $\mathcal{D}_u$ to obtain a high-performance semantic segmentation model.
	
	Fig. \ref{fw} presents the overall architecture and the training procedure of our method, which is built from a basic semi-supervised model consisting of two parts. One is a CNN-based encoder that extracts high-dimensional representations for given input images. The other is a decoders with a classification head and a feature embedding head respectively.  The classifier outputs label prediction vectors. 
	The embedding head constructs an embedding space by mapping high-dimensional representations to  low-dimensional ones. 
	
	We need to train labeled and unlabeled data separately, and the complete loss is shown as follows
	\begin{align}
		\mathcal{L}_{total}=\mathcal{L}_{sup}(X_l,Y_l)+\lambda_{unsup}\mathcal{L}_{unsup}(X_u),
	\end{align}
	where $\mathcal{L}_{sup}$ and $\mathcal{L}_{unsup}$ denote supervised loss and unsupervised loss, respectively. $\lambda_{unsup}$ denotes the weight of unsupervised loss. 
	
	For labeled images, we apply typical pixel-wise cross-entropy loss at the segmentation head
	\begin{align}
		\mathcal{L}_{sup}=\sum_{i,j}\sum_{c}Y_l^{(i,j,c)}\log{(P^{(i,j,c)})},
	\end{align}
	where $Y_l$ is ground truth and $P\in{\mathbb{R}^{H\times{W}\times{C}}}$ is prediction vector. 
	
	For unlabeled images, we perform a strong augmentation and a weak one on a single unlabeled image. Generally, it is necessary to perform consistency loss for the two kinds of augmented data. Traditionally, unsupervised loss $\mathcal{L}_{unsup}$ is written as follows
	\begin{align}
		\sum_{i,j}\sum_{c}\hat{Y}^{(i,j,c)}\log{(P^{(i,j,c)})},
	\end{align}
	where $\hat{Y}$ is pseudo-labels, while in this work, we propose a novel unsupervised loss based on multi-level graphs to correct erroneous pseudo-labels.

	\subsection{Semantic-Level Graph and Class-Level Graph Construction}
	\textbf{Semantic-Level Graph (SLG)} preserves semantic relationship among features. Noisy pseudo-labels can be efficiently calibrated by aggregating semantic neighborhood knowledge. Specifically, we construct an undirected 
	graph $\mathcal{G}^{(S)}=(\mathcal{V}^{(S)},\mathcal{E}^{(S)}, \mathcal{X}^{(S)})$ from all unlabeled pixels in a mini-batch, where $\mathcal{V}^{(S)}$ represents the set of vertices corresponding features 
	generated from the embedding head. Variable $\mathcal{E}^{(S)}$ represents the set of edges, which measures the affinity between pairs of features at pixel-level. $\mathcal{X}^{(S)}$ is the feature matrix and each row corresponds to a node feature.  An intuitive way to construct edges is to connect all pixels, but this method yields expensive computation complexity and some original feature may be covered by incorrect neighbors. Therefore, we only keep $k$ nearest neighbors and use an 
	affinity matrix $\hat{\mathcal{A}}\in{\mathbb{R}^{n\times{n}}}$ to model such relationship
	\begin{align}
		\label{g_s}
		\hat{\mathcal{A}}_{i j} & = \left\{\begin{array}{cc}
			\!\!\!{\left[\frac{\mathcal{X}^{(S)}_{i,:}\cdot\mathcal{X}^{(S)}_{j,:}}{\left\|\mathcal{X}^{(S)}_{i,:}\right\| \times\left\|\mathcal{X}^{(S)}_{j,:}\right\|}\right]_{+}^{\gamma},} \!\!\!\!\!\! & \text { if } i \neq j \text { and } j \in \mathrm{NN}_{k}\left(\mathcal{X}^{(S)}_{i,:}\right), \\
			0, & \text { otherwise },
		\end{array}\right.
	\end{align}
	where $\mathcal{X}^{(S)}_{i,:}$ is a feature vector corresponding to pixel $i$, $\gamma$ is a scaling parameter and we set it to $1$, $\mathrm{NN}_k(\mathcal{X}^{(S)}_{i,:})$ is the set of $k$ nearest neighbors of $\mathcal{X}^{(S)}_{j,:}$, and $n=H\times{W}$ is the number of pixels in each image,  $\left[ \ \cdot \   \right]_{+}=\text{max}(0,\cdot)$. Then, we normalize the obtained affinity matrix 
	\begin{align}
		\label{g_s_n}
		\mathcal{A}  = D^{-\frac{1}{2}}\left(\hat{\mathcal{A}}+\hat{\mathcal{A}}^T \right) D^{-\frac{1}{2}},
	\end{align}
	where $D\in{\mathbb{R}^{n\times{n}}}$ is the diagonal degree matrix of $ \mathcal{A}$ with $D_{ii}=\sum_j{\left(\hat{\mathcal{A}}_{ij}+\hat{\mathcal{A}}_{ji}^T \right)}$. We use the summation $\hat{\mathcal{A}}_{ij}+\hat{\mathcal{A}}_{ji}^T $ to keep symmetricity.  
	
	\textbf {Class-Level Graph (CLG)} preserves class consistency information, namely, edges between nodes indicating pairs of pixels belonging to the same class. By comparing the similarity of predicted class labels, relationship between feature nodes belonging to the same class can be strengthened. In addition, this aggregation can endow the node features with context-aware ability and obtain more accurate feature representation. Analogous to SLG, specifically, we construct an undirected graph $\mathcal{G}^{(C)}=(\mathcal{V}^{(C)},\mathcal{E}^{(C)}, \mathcal{X}^{(C)})$
	from all unlabeled pixels in a mini-batch, where $\mathcal{V}^{(C)}$ denotes the set of vertices corresponding to the output labels 
	from the classification head. $\mathcal{E}^{(C)}$ represents the set of edges, which measures affinity between nodes. 
	$\mathcal{X}^{(C)}$ is confidence matrix and each row corresponding to the predicted probability vector.
	We use another affinity matrix $\hat{\mathcal{W}}\in{\mathbb{R}^{n\times{n}}}$ to model the relationship between the label confidence scores:
	\begin{align}
		\label{g_l}
		\hat{\mathcal{W}}_{i j}  = \left\{\begin{array}{ll}
			1 & \text { if } i  = j, \\
			&\\
			\mathcal{X}^{(C)}_{i,:}\cdot \mathcal{X}^{(C)}_{j,:} & \text { if  $i\ne j$  and }\\ &\text { $\underset{i}{\arg\max{\mathcal{X}^{(C)}_{i,:}}}  = \underset{j}{\arg\max{\mathcal{X}^{(C)}_{j,:}}}$},  \\
			0 & \text { otherwise, }
		\end{array}\right.
	\end{align}
	where $\mathcal{X}^{(C)}_{i,:}$ is  a predicted probability vector corresponding to pixel $i$. 
	Thus, pixels that do not belong to the same class in terms of pseudo-labels are not connected, and each pixel is connected to itself with weight one. 
	Then, we normalize the obtained similarity matrix
	\begin{align}
		\label{g_l_n}
		\mathcal{W}  = \hat{D}^{-\frac{1}{2}}\hat{\mathcal{W}} \hat{D}^{-\frac{1}{2}},
	\end{align}
	where $\hat{D}$ is the diagonal degree matrix of $\hat{\mathcal{W}}$ with $\hat{D}_{ii}=\sum_j{\hat{\mathcal{W}}_{ij}}$. {Because $\hat{\mathcal{W}}$ is a symmetric matrix by definition, we do not need to use the summation of it with its transpose.} 

	\subsection{Self-Distillation Between Semantic-Level Graph and Class-Level Graph}
	In this section, we elaborate self-distillation process of multi-level graphs. 
	During each semi-supervised iteration, we need to perform $K$ times of graph convolution operations on the constructed graph. In the $k$th graph convolution operation, 
	
	we first update Class-Level Graph node $\hat{\mathcal{V}}^{(C,k)}$ with the following propagation rule:
	\begin{align}
		\hat{\mathcal{V}}_{i,:}^{(C,k)}  = f_{\text { C }}^{(k)}
		\left(\left[\mathcal{X}^{(C,k-1)}_{i,:}, 
		\sum_{j  = 1}^{n} \mathcal{A}_{ij}^{(k)} \cdot \mathcal{X}^{(C,k-1)}_{j,:}\right];\theta_{k}^{(L)}\right),\label{eq5}
	\end{align}
	where $f_{\text {C}}^{(k)}$ is a non-linear function parameterized by $\theta_{k}^{(L)}$, and $[\cdot,\cdot]$ is a concatenation function. {By this operation, nearby pseudo-labels are concatenated according to $\mathcal{A}^{(k)}$ and are then embedded in a new space, which can progressively rectify  noisy pseudo-labels.} Regions with low confidence in the prediction are more probably to be noisy, so we should focus more on low confidence regions. Following Flexmatch, we use a class-specific dynamic threshold to separate high-confidence and low-confidence labels. The class-specific dynamic thresholds are denoted as follows:
	\begin{align}
		\eta^{(t,c)} = \frac{\delta ^{(t,c)} }
		{ \underset{c}{\max \delta}}\cdot\sigma,
	\end{align}
	where $\sigma$ is a fixed threshold, and $\delta^{(t,c)}=\sum_{i = 1}^n\mathbbm{1}(\max p_i>\sigma)\cdot \mathbbm{1}(\arg\max p_i = c)$ indicates learning effect of current class $c$ in the current mini-batch $t$ , $p_i$ is prediction vector at original segmentation head. $\mathbbm{1}$ is a function that returns one when its input is true and returns zero otherwise. According to the dynamic threshold, we update Class-Level Graph node $\mathcal{V}^{(C,k)}$ using the following rule:
	\begin{align}
		\mathcal{V}_{i,:}^{(C,k)}=&[\alpha \hat{\mathcal{V}}_{i,:}^{(C,k)}+(1\!-\!\alpha) \mathcal{V}_{i,:}^{(C,k-1)}]\!\!\cdot\!\!\mathbbm{1}(\max p_i\ge \eta^{(t,c)}) \nonumber \\
		&+[(1\!-\!\alpha) \hat{\mathcal{V}}_{i,:}^{(C,k)}+\alpha \mathcal{V}_{i,:}^{(C,k-1)}]\!\!\cdot\!\!\mathbbm{1}(\max p_i< \eta^{(t,c)}),
	\end{align}
	where $\alpha$ is a hyperparameter that controls the proportion of nodes before and after label propagation. {This equation means to apply the correction according to the confidences of transformed pseudo-labels before and after rectification, represented by $\mathcal{V}$ and $\hat{\mathcal{V}}$ respectively.} After updating the nodes, we update edge connection $\mathcal{W}^{(k)}$ by Eq.(\ref{g_l}).
	
	After $\mathcal{G}^{(C)}$ update, the semantic-level graph $\mathcal{G}^{(S)}$ node is reconstructed by utilizing the information in $\mathcal{G}^{(C)}$. Node features in $\mathcal{G}^{(S)}$ through $\mathcal{W}^{(k)}$ aggregates correct neighbors and update process is shown as follows:
	\begin{align}
		\mathcal{V}_{i,:}^{(S,k)}  = f_{\text {S }}^{(k)}\left(\left[ \mathcal{X}^{(S,k-1)}_{i,:}, \sum_{j  = 1}^{n} \mathcal{W}_{ij}^{(k)} \cdot  \mathcal{X}^{(S,k-1)}_{j,:}\right] ; \theta_{k}^{(S)}\right),
	\end{align}
	where $f_{\text {S }}^{(k)}$ is a non-linear function parameterized by $\theta_{k}^{(S)}$. {By this operation, features from the same category are concatenated according to $\mathcal{W}^{(k)}$ and are then embedded into a new space, which can generate more representative features.} After updating the nodes, we update edge connection $\mathcal{A}^{(k)}$ by Eq.(\ref{g_s}). The complete multi-level graph algorithm is shown in Algorithm \ref{alg:algorithm}.
	
	\subsection{Training Process}
	We need to train the SLG branch and CLG branch respectively. 
	
	\textbf{SLG Loss.} For training SLG, we propose a contrastive learning scheme that incorporates both intra-image and inter-image comparison. The loss is shown as follows: 
	\begin{align}
		\label{iccl_loss}
		\mathcal{L}_{SLG}^{(S,k)}= &\lambda \sum_{i,j} [Y_{\Pi}^{i,j}
		(\tilde{v}_i^k\cdot \tilde{v}_j^k-1)^2+(1-Y_{\Pi}^{i,j})(\tilde{v}_i^k\cdot \tilde{v}_j^k)^2] \nonumber \\
		&-(1-\lambda )\sum_{i}\log\frac{\exp{(\tilde{v}_i^k\cdot p_{\tilde{y}_i}/\tau)}}
		{\sum_{p_c^k \in \mathcal{P}^k }\exp{(\tilde{v}_i^k\cdot p_{c}/\tau)}}
	\end{align}
	where $\lambda$ is a weight factor that controls balance of the two loss items. $Y_{\Pi}^{i,j}\in \{0,1\}^{n\times{n}}$ is $1$ if $\tilde{v}_i^k$ and $\tilde{v}_i^k$ have same label and 0 otherwise, $\tilde{v}_i^k$ represent features in $k$th SLG. Tildes above vectors represent strong augmentations. $\mathcal{P}^k=\{p_c^k\}_{c=1}^C$ denotes prototypes. $\tilde{y}_i\in{[1,2,\cdots,C]}$ is pseudo-label at pixel $i$ obtained by weakly augmented version, which determines positive prototype $p_{\tilde{y}_i}$. $\tau$ is a temperature, we set it to $0.1$. The first term works within a single image and, intuitively, it enforces the network to learn highly confident semantic affinities. The second term evaluates the difference between node features and prototypes to facilitate movement of features toward the correct clustering center. The prototypes $\mathcal{P}^k$ aggregate same class features of all data, so it has cross-image knowledge. We use feature averaging to calculate the prototypes, and the obtained prototypes can be regarded as the feature centers of each class. The prototypes are calculated as follows
	\begin{align}
		p_c^{(k,t)}=\frac{\sum_i v_i^k\cdot{\mathbbm{1}(y_i=c)}}{\sum_i \mathbbm{1}(y_i=c) },
	\end{align}
	where $v_i^k$ is feature in $k$th SLG, and $y_i$ is ground truth at position $i$ for labeled images. However, when a few examples are used to represent the category center, there is a deviation. Therefore, for unlabeled images, we use the pseudo-label $\tilde{y}_i$ instead of $y_i$ to compute the prototype. To solve the problem that the prototypes representation can be imbalanced when using class-imbalanced data during training, we estimate the prototypes as exponential moving averages of small batches of cluster masses so that we can keep track of the slow-moving prototypes. Specifically, in each iteration, the prototypes are estimated as follows:
	\begin{align}
		p_c^{(k,t)} \gets \beta p_c^{(k,t)}+(1-\beta)p_c^{(k,t-1)},
	\end{align}
	where $\beta$ is a hyperparameter that controls the speed of prototype updates, we set it to $0.99$.
	
	\textbf{CLG Loss.} For the CLG training, we propose a dynamic weighted cross-entropy loss to further reduce the effect of noisy labels. The loss is shown as follows:
	\begin{align}
		\label{dwcl_loss}
		\mathcal{L}_{CLG}^{(L,k)}=\sum_{i,j}\sum_{c}\omega_{i,j}^k\hat{Y}^{(i,j,c)}\log{(P_k^{(i,j,c)})},
	\end{align}
	where $\hat{Y}$ is the pseudo-labels and $P_k$ is the output of the $k$th CLG confidence matrix, i.e. $\mathcal{X}^{(C,k)}$. $\hat{Y}=\arg\max\sum_k{\mathcal{X}^{(C,k)}}$ by aggregating the confidence matirx of $K$ CLGs to obtain more accurate output. The dynamic weight is denoted as follows:
	\begin{align}
		\omega_{i,j}^k=P_k^{(i,j)}[\hat{Y}^{(i,j)}],
	\end{align}
	we use confidence from strongly augmented data as feedback instead of confidence from weakly augmented data.
	
	\textbf{Total Loss For Unsupervised Learning.} 
	In summary, the total unlabeled loss at each mini-batch is as follows:
	\begin{align}
		\mathcal{L}_{unsup}=\lambda_{SLG}\sum_k{\mathcal{L}_{SLG}^{(S,k)}}+\lambda_{CLG}\sum_k \mathcal{L}_{CLG}^{(L,k)},
	\end{align}
	where $\lambda_{SLG}$ denotes the weight of SLG loss and $\lambda_{CLG}$ denotes the weight of CLG loss. We use lose for each graph layer to accelerate training.

	\begin{algorithm}[!th]
		\begin{minipage}{0.9\linewidth}
			\caption{The Alternating Update Process of SLG and CLG}
			\label{alg:algorithm}
			\textbf{Input}: Initial  $\mathcal{G}_{init}^{(S)}=\{\mathcal{V}_{init}^{(S)},\mathcal{E}_{init}^{(S)},\mathcal{X}_{init}^{(S)}\}$, Initial $\mathcal{G}_{init}^C=\{\mathcal{V}_{init}^{(C)},\mathcal{E}_{init}^{(C)},\mathcal{X}_{init}^{(C)}\}$\\
			\textbf{Output}: Refined SLG and refined CLG 
			\begin{algorithmic}[1]
				\For{$k$ in  $0:K$ do}
				\State {\color{blue}\#Stage \uppercase\expandafter{\romannumeral1}: Compute the semantic affinity matrix of SLG}
				\State Update $\mathcal{A}^{(k)} $ by Eq. (\ref{g_s_n});
				\State {\color{blue}\#Stage \uppercase\expandafter{\romannumeral2}: Update the nodes of the CLG}
				\State Update $\mathcal{V}^{(C,k)} $ by Eq. (10);
				\State {\color{blue}\#Stage \uppercase\expandafter{\romannumeral3}: Compute the class consistency matrix of CLG}
				\State Update $\mathcal{W}^{(k)} $ by Eq. (\ref{g_l_n});
				\State {\color{blue}\#Stage \uppercase\expandafter{\romannumeral4}: Update the nodes of the SLG}
				\State Update $\mathcal{V}^{(S,k)} $ by Eq. (11);

				\EndFor

			\end{algorithmic}
		\end{minipage}
	\end{algorithm}

	\section{Experimental Results}

	\subsection{Experiment Setting}
	\textbf{Datasets.} 
	We conduct experiments on benchmark datasets: Cityscapes \cite{cordts2016cityscapes} and PASCAL VOC 2012 \cite{everingham2015pascal}. Cityscapes is a real-world urban traffic scenes with image size $1024\times2048$ and each pixel belongs to one of nineteen classes. With this data set, we use 2975 images for training, 500 images for validation. PASCAL VOC 2012 is a natural landscape dataset with twenty-one classes including a background class. With this dataset, we use 10582 images for training and 1449 images for validation. We create different partition protocols for the above datasets, such as 1/30, 1/16, 1/8 and 1/4 partition protocols.

	\begin{table}[t]
		
		\caption{Comparison with state-of-the-art methods on Cityscapes val set under different partition protocols. All methods use DeepLabV2 with ResNet101. 
		}
		\centering
		\label{cityresultshow}

		\resizebox{\linewidth}{!}{
			\begin{tabular}{l|cccc}  			
				\toprule   
				\multirow{2}{*}{\textbf{Methods}} & \multicolumn{4}{c}{\textbf{Cityscapes, DeepLabV2}}  \\
				\cmidrule{2-5}
				& 1/30 (100) & 1/8 (372) & 1/4 (744) & 1/2 (1487) \\  
				\midrule
				Sup. Base.  &  48.36  &  57.80  &  61.63  &  63.13  \\
				\midrule
				AdvSemiSeg \cite{hung2018adversarial}  &  -  &  58.80 (+1.00) &  62.30 (+0.67) &  65.70 (+2.57)  \\
				s4GAN \cite{mittal2019semi}  &  -  &  59.30 (+1.50) &  61.90 (+0.27) &  -  \\
				French \emph{et al.} \cite{french2019semi}  &  51.20 (+2.84) &  60.34 (+2.54) &  63.87 (+2.24) &  -  \\
				DST-CBC \cite{feng2020semi}  &  48.70 (+0.34) &  60.50 (+2.7) &  64.40 (+2.77) &  -  \\
				ClassMix \cite{olsson2021classmix}  &  54.07 (+5.71) &  61.35 (+3.55) &  63.63 (+2.00) &  66.29 (+3.16) \\
				DMT \cite{feng2020dmt}  &  54.80 (+6.44) &  63.03 (+5.23) &  -   &  -  \\
				ECS \cite{mendel2020semi}  &  -  &  60.26 (+2.46) &  63.77 (+2.14) &  -   \\
				$\text{C}^3$-SemiSeg \cite{DBLP:conf/iccv/0001X0GH21}  &  55.17 (+6.81) &  63.23 (+5.43) &  65.50 (+3.87) &  -   \\
				Contra-SemiSeg \cite{alonso2021semi}  &  59.40 (+11.04) &  64.40 (+6.60) &  65.92 (+4.29) &  -  \\
				CARD \cite{wang2022card}  &  55.63 (+7.27) &  65.10 (+7.30) &  66.45 (+4.82) &  -  \\
				\midrule
				\textbf{MLLC}   &  \textbf{59.61 {(+11.25)}} &  \textbf{65.90 {(+8.10)}} &  \textbf{67.42  {(+5.79)}} &  \textbf{68.73 {(+5.60)}} \\
				\bottomrule  
			\end{tabular}
		}
		
	\end{table}

	\begin{table}[t]
		
		\caption{Comparison with state-of-the-art methods on Cityscapes val set under different partition protocols. All methods use DeepLabV3+ with ResNet101. 
		}
		\centering
		\label{cityresultshowv3}

		\resizebox{\linewidth}{!}{
			\begin{tabular}{l|cccc}  			
				\toprule   
				\multirow{2}{*}{\textbf{Methods}} & \multicolumn{4}{c}{\textbf{Cityscapes, DeepLabV3+}}  \\
				\cmidrule{2-5}
				& 1/16 (186) & 1/8 (372) & 1/4 (744) & 1/2 (1487) \\  
				\midrule
				Sup. Base.  &  66.27 &  71.63  &  75.36  &  78.31  \\
				\midrule
				MT \cite{tarvainen2017mean} & 68.08(+1.81) & 73.71(+2.08) & 76.56(+1.20) & 78.59(+0.28) \\
				CCT \cite{ouali2020semi} & 69.64(+3.37) & 74.48(+2.85) & 76.35(+0.99) & 78.59(+0.28) \\
				GCT \cite{ke2020guided} & 66.9(+0.63) & 72.96(+1.33) & 76.45(+1.09) & 78.58(+0.27) \\
				CPS \cite{chen2021semi} & 70.5(+4.23) & 75.71(+4.08) & 77.41(+2.05) & 80.08(+1.77) \\
				PS-MT \cite{liu2022perturbed} & - & 76.89(+5.26) & 77.6(+2.24) & 79.09(+0.78) \\
				$\text{U}^{2}$PL \cite{wang2022semi}& 70.30(+4.03) & 74.37(+2.74) & 76.47(+1.11) & 79.05(+0.74) \\
				PCR \cite{xu2022semi} & 73.41(+7.14) & 76.31(+4.68) & 78.4(+3.04) & 79.11(+0.8) \\
				AEL \cite{hu2021semi} & 74.45(+8.18) & 75.55(+3.92) & 77.48(+2.12) & 79.01(+0.7) \\
				GTA-Seg \cite{jin2022semi} & 69.38(+3.11) & 72.02(+0.39) & 76.08(+0.72) & - \\
				CISC-R \cite{wu2023querying} & - & 75.03(+3.40) & 77.02(+1.66) & - \\
				UniMatch \cite{yang2023revisiting} & \textbf{76.60(+10.33)} & 77.90(+6.27) & 79.20(+3.84) & 79.50(+1.19) \\
				FPL \cite{qiao2023fuzzy} & 73.20(+6.93) & 75.74(+4.11) & 78.19(+2.83) & 79.19(+0.88) \\
				DGCL \cite{wang2023hunting} & 73.18(+6.89) & 77.29(+5.66) & 78.48(+3.12) & \textbf{80.71(+2.40)} \\
				CCVC \cite{wang2023conflict} & 74.90(+8.63) & 76.40(+4.77) & 77.30(+1.94) & - \\
				PCR \cite{xu2022semi} & 73.41(+7.14) & 76.31(+4.68) & 78.40(+3.04) & 79.11(0.80) \\
				\midrule
				\textbf{MLLC}   &  75.43 {(+9.16)} &  \textbf{77.92 {(+6.29)}} &  \textbf{79.31  {(+3.95)}} &  80.22 {(+1.91)} \\
				\bottomrule  
			\end{tabular}
		}
		
	\end{table}

	\textbf{Evaluation.}
	We employ DeepLabV2 \cite{chen2017deeplab} or DeepLabV3+ \cite{chen2018encoder}, which are pre-trained on ImageNet \cite{deng2009imagenet}, as our segmentation models to ensure a fair comparison with previous work. Both of the classification head and embedding head consists of a \textbf{Conv-BN-ReLU} blocks. Both heads map output into given categories and 256-dimensional feature space, respectively. Following previous work, we use the mean Intersection-over-Union (mIoU) metric to evaluate the segmentation performance of all datasets. In semantic segmentation, mIoU represents the ratio of intersection and union of ground truth and predicted labels. mIoU ranges from 0 to 1, and higher values indicate better performance of segmentation model.

	\textbf{Implementation Details.} 
	All experiments use the stochastic gradient descent (SGD) optimizer to train Cityscapes and VOC for 100 and 150 epochs with batch size $16$ under a polynomial decay strategy $1-(\frac{\text{iter}}{\text{total\_iter}})^{0.9}$. The trade-off weights $\lambda_{contr}$ and $\lambda_{dwc}$ are set to $0.1$ and $1$, respectively. Following \cite{zhang2021flexmatch}, we set $\sigma$ to $0.95$. The label propagation control hyperparameter $\alpha$ is set to $0.8$. The balance parameter $\lambda$ of SLG Loss is set to $0.5$. The number of loops in refining multi-level graphs, $K$, is set to two. In experiments, we use various data augmentations, where the weak data augmentations used in the teacher network include random cropping, random scaling and random flipping, etc., and the strong data augmentations used in the student network include color jitter, blur, and CutMix, etc. Following \cite{feng2020dmt}, in experiments with deeplabv2 as the backbone, Cityscapes images are resized to 512$\times$1024 without random scaling and flipping, and then the images are randomly cropped to the size of 256$\times$512. In PASCAL VOC 2012, random scaling and flipping are used with a scaling factor of 0.5 to 1.5 and cropping size 321$\times$321. Following \cite{wang2022semi}, in experiments with deeplabv3+ as the backbone, we use crop sizes 512$\times$512 and 800$\times$800 on PASCAL VOC 2012 and Cityscapes, respectively.
	
	\subsection{Comparison with State-of-the-Art Methods}
	In this section, we compare MLLC with other frameworks with the same backbone network and the same settings to ensure the fairness of the comparison.

	{
	\textbf{Cityscapes.} 
	Table \ref{cityresultshow} shows the results of our method compared with other state-of-the-art methods using DeepLabV2 on the Cityscapes. Specifically, MLLC significantly outperforms the supervised baseline (Sup. Base.) by 11.25\%, 8.10\%, 5.79\%, and 5.60\% under 1/30, 1/8, 1/4, and 1/2 partition protocols, respectively. It also outperforms all the competitors, obtaining 59.61\%, 65.90\%, 67.42\%, and 68.73\% performance in partitioning protocols 1/30, 1/8, 1/4, and 1/2, respectively. 
	
	We can also observe that when less labeled data is available, e.g., 100 and 372 labels, MLLC has a significant performance improvement compared to supervised baseline (trained using only labeled data). The brackets in the table show the performance difference between the different methods and supervised baseline. Larger numbers indicate that the model has a higher improvement over supervised baseline. In addition, we also observe that the gap between the semi-supervised semantic segmentation method and supervised baseline keeps decreasing as the proportion of labeled data increases, but our method can still improve by 5.79\% and 5.60\& using 1/4 fine label and 1/2 fine label, respectively. 
	
	In summary, the results of the Cityscapes dataset show significant improvements in MLLC under various partitioning protocols, and the method can effectively utilize both labeled and unlabeled data. In Table \ref{cityresultshowv3}, shows the results of our method compared with other state-of-the-art methods using DeepLabV3+ on the Cityscapes. Specifically, MLLC significantly outperforms the supervised baseline (Sup. Base.) by 9.16\%, 6.29\%, 3.95\%, and 1.91\% under 1/16, 1/8, 1/4, and 1/2 partition protocols, respectively.

	\begin{table}[t]
		
		\caption{Comparison with state-of-the-art methods on PASCAL VOC 2012
			val set under different partition protocols. All methods use DeepLabV2 with ResNet101.}
		\centering
		\label{vocresultshow}

		\resizebox{\linewidth}{!}{
			
			\begin{tabular}{l|cccc}

				\toprule   
				\multirow{2}{*}{\textbf{Methods}} & \multicolumn{4}{c}{\textbf{PASCAL VOC 2012, DeepLabV2}}  \\
				\cmidrule{2-5}
				
				& 1/50 (212) & 1/16 (529) & 1/8 (1322) & 1/4 (2645) \\  
				\midrule
				
				Sup. Base.  &  55.69  &  62.81  &  67.15  &  70.21  \\
				\midrule
				AdvSemiSeg \cite{hung2018adversarial}  &  57.20 (+1.51) &  65.67 (+2.86) &  69.50 (+2.35) &  72.10 (+1.89)  \\
				s4GAN \cite{mittal2019semi}  &  63.30 (+7.61) &  -  &  71.40 (+4.25) &  -   \\
				French \emph{et al.} \cite{french2019semi}  &  64.81 (+9.12) &  -  &  67.60 (+0.45) &  -  \\
				DST-CBC \cite{feng2020semi}  &  65.50 (+9.81) &  -  &  70.70 (+3.55) &  71.80 (+1.59) \\
				MT \cite{tarvainen2017mean}  &  -  &  66.08 (+3.27) &  69.81 (+2.66) &  71.28 (+1.07) \\
				ClassMix \cite{olsson2021classmix}  &  66.15 (+10.46) &  -  &  71.00 (+3.85) &  72.45 (+2.24)\\
				DMT \cite{feng2020dmt}  &  67.15 (+11.46) &  -  &  72.7  (+5.55) &  -  \\
				ECS \cite{mendel2020semi}  &  -  &  -  &  -  &  72.60 (+2.39) \\
				GCT \cite{ke2020guided}  &  -  &  67.19 (+4.38) &  -  &  73.62 (+3.41) \\
				Contra-SemiSeg \cite{alonso2021semi}  &  67.90 (+12.21) &  -  &  71.6 (+4.45) &  -  \\
				CARD \cite{wang2022card}  &  70.94 (+15.25) &  -  &  74.07 (+6.92) &  -  \\
				\midrule
				\textbf{MLLC}   &  \textbf{71.18  {(+15.49)}} &  \textbf{72.13  {(+9.32)}} &  \textbf{74.12   {(+6.97)}} &  \textbf{75.92  {(+5.71)}}\\
				
				\bottomrule  
			\end{tabular}
		}
		
	\end{table}
	
	\begin{table}[t]
		
		\caption{Comparison with state-of-the-art methods on PASCAL VOC 2012
			val set under different partition protocols. All methods use DeepLabV3+ with ResNet101.}
		\centering
		\label{vocresultshowv3}

		\resizebox{\linewidth}{!}{
			
			\begin{tabular}{l|cccc}

				\toprule   
				\multirow{2}{*}{\textbf{Methods}} & \multicolumn{4}{c}{\textbf{PASCAL VOC 2012, DeepLabV3+}}  \\
				\cmidrule{2-5}
				
				& 1/16 (662) & 1/8 (1323) & 1/4 (2646) & 1/2 (5291) \\  
				\midrule
				
				Sup. Base.  &  67.55  &  72.91  &  75.12  &  77.10  \\
				\midrule
				MT \cite{tarvainen2017mean} & 70.51(+2.96) & 71.53(-1.38) & 73.02(-2.1) & 76.58(-0.52) \\
				CutMix \cite{mittal2019semi} & 71.66(+4.11) & 75.51(+2.6) & 77.33(+2.21) & 78.21(+1.11) \\
				CCT \cite{ouali2020semi} & 71.86(+4.31) & 73.68(+0.77) & 76.51(+1.39) & 77.40(+0.30) \\
				GCT \cite{ke2020guided} & 70.9(+3.35) & 73.29(+0.38) & 76.66(+1.54) & 77.98(+0.88) \\
				CPS \cite{chen2021semi} & 74.48(+6.93) & 76.44(+3.53) & 77.68(+2.56) & 78.64(+1.54) \\
				AEL \cite{hu2021semi}  & 77.2(+9.65) & 77.57(+4.66) & 78.06(+2.94) & 80.29(+3.19) \\
				$\text{U}^{2}$PL \cite{wang2022semi} & 77.21(+9.66) & 79.01(+6.10) & 79.3(+4.18) & 80.50(+3.40) \\
				CPCL \cite{fan2023conservative} & 73.44(+5.89) & 76.40(+3.49) & 77.16(+2.04) & 77.67(+0.57) \\
				Unimatch \cite{yang2023revisiting} & 78.10(+10.55) & 78.40(+5.49) & 79.20(+2.10) & - \\
				FPL \cite{qiao2023fuzzy} & 74.98(+7.43) & 76.73(+3.82) & 78.35(+3.23) & - \\
				DGCL \cite{wang2023hunting} & 76.61(+9.06) & 78.37(+5.46) & 79.31(+4.19) & 80.96(+3.86) \\
				PCR \cite{xu2022semi} & 78.60(+11.05) & \textbf{80.71(+7.80)} & 80.78(+5.66) & 80.91(+3.81) \\
				CCVC \cite{wang2023conflict} & 77.20(+9.65) & 78.40(+5.49) & 79.00(+3.99) & - \\
				AugSeg \cite{zhao2023augmentation} & 77.01(+9.46) & 77.31(+4.40) & 78.82(+3.68) & - \\

				\midrule
				\textbf{MLLC}   &  \textbf{78.93  {(+11.38)}} &  80.29  {(+7.38)} &  \textbf{80.81   {(+5.69)}} &  \textbf{80.96  {(+3.86)}}\\
				
				\bottomrule  
			\end{tabular}
		}
		
	\end{table}
	
	\textbf{PASCAL VOC 2012.}
	Table \ref{vocresultshow} shows the results of MLLC compared with other methods using DeepLabv2 on PASCAL VOC 2012. Our approach significantly outperforms the supervised baseline, increasing performance by 15.49\%, 9.32\%, 6.97\%, and 5.71\% with 1/50, 1/16, 1/8, and 1/4 labeled data, respectively. MLLC also achieves state-of-the-art performance, especially in case of very rare labels, obtaining 71.18\%, 72.13\%, 74.12\%, and 75.92\% performance in partitioning protocols 1/50, 1/16, 1/8, and 1/4, respectively. 
	
	In Table \ref{vocresultshowv3}, shows the results of MLLC compared with other methods using DeepLabv3+ on PASCAL VOC 2012. Specifically, MLLC significantly outperforms the supervised baseline (Sup. Base.) by 11.38\%, 7.38\%, 5.69\%, and 3.86\% under 1/16, 1/8, 1/4, and 1/2 partition protocols, respectively.
	
	When comparing using DeepLabV3+ as shown in Table \ref{vocresultshowv3}, partitioning protocols and experimental settings follow PseudoSeg\cite{zou2020pseudoseg}. PseudoSeg randomly selects 1/2, 1/4, 1/8 and 1/16 of the A training set (about 1400 images) as labeled data, and rest of data and enhancement set (about 9000 images) as unlabeled data. Our approach outperforms the supervised baseline by 29.41\%, 23.52\%, 16.99\%, and 13.02\%  with 1/16, 1/8, 1/4, and 1/2 labeled data, respectively. MLLC also achieves state-of-the-art performance, obtaining 70.63\%, 74.34\%, 77.36\%, and 79.31\% performance in partitioning protocols 1/16, 1/8, 1/4, and 1/2, respectively. Note that in low-label data, e.g., 92 and 183, MLLC outperforms the performance of state-of-the-art methods 2.65\% and 3.31\%, respectively. In all these experiments, MLLC is superior to other methods.	
	{
	Table. \ref{trf} illustrates a comparison between a classic backbone and a more powerful transformer-based backbone \cite{xie2021segformer,hoyer2022daformer}, and the results show that MLLC can significantly improve the performance of segmentation for both backbone models.}

	\begin{table}[t]
		
		\setlength\tabcolsep{1pt}  
		\caption{Comparison with state-of-the-art methods on PASCAL VOC 2012 val set under different partition protocols in low-data regime. All methods use DeepLabV3+ with ResNet101. 
		}
		\centering
		\tabcolsep=6pt
		\label{vocresultshowv3}

		\resizebox{\linewidth}{!}{
			\begin{tabular}{l|cccc}  
				
				\toprule   
				\multirow{2}{*}{\textbf{Methods}} & \multicolumn{4}{c}{\textbf{PASCAL VOC 2012 dataset in low-data regime}}  \\
				\cmidrule{2-5}
				& 1/16 (92) & 1/8 (183) & 1/4 (366) & 1/2 (732) \\  
				\midrule
				Sup. Base.  &  41.22  &  50.82  &  60.37  &  66.29 \\
				\midrule
				AdvSemSeg \cite{hung2018adversarial}  &  39.69 (-1.53) &  47.58 (-3.24) &  59.97 (-0.4) &  65.27 (-1.02)\\
				CCT \cite{ouali2020semi}  &  33.10 (-8.12) &  47.60 (-3.22) &  58.80 (-1.57) &  62.10 (-4.19)\\
				GCT \cite{ke2020guided}  &  46.04 (+4.82) &  54.98 (+4.16) &  64.71 (+4.34) &  70.67 (+4.38)\\
				VAT \cite{DBLP:journals/pami/MiyatoMKI19}  &  36.92 (-4.30) &  49.35 (-1.47) &  56.88 (-3.49) &  63.34 (-2.95)\\
				CutMix \cite{french2019semi}  &  55.58 (+14.36) &  63.20 (+12.38) &  68.36 (+7.99) &  69.84 (+3.55)\\
				PseudoSeg \cite{zou2020pseudoseg}  &  57.60 (+16.38) &  65.50 (+14.68) &  69.14 (+8.77) &  72.41 (+6.12)\\
				P$\text{C}^2$Seg \cite{DBLP:conf/iccv/ZhongYWY0W21}  &  57.00 (+15.78) &  66.28 (+15.46) &  69.78 (+9.41) &  73.05 (+6.76)\\
				CPS \cite{chen2021semi} &  64.07 (+22.85) &  67.42 (+16.60) &  71.71 (+11.34) &  75.88(+9.59)\\
				ST++ \cite{yang2022st++} &  65.23 (+24.01) &  71.01 (+20.19) &  74.59 (+14.22) &  77.33(+11.04)\\
				$\text{U}^2$PL \cite{wang2022semi} &  67.98 (+26.76) &  69.15 (+18.33) &  73.66 (+13.29) &  76.16(+9.87)\\
				PS-MT \cite{liu2022perturbed} &  65.80 (+24.58) &  69.58 (+18.76) &  76.57 (+16.20) &  78.42 (+12.13)\\
				\midrule
				\textbf{MLLC}   &  \textbf{70.63 {(+29.41)}} &  \textbf{74.34 {(+23.52)}} &  \textbf{77.36  {(+16.99)}} &  \textbf{79.31{(+13.02)}}\\
				
				\bottomrule  
			\end{tabular}
		}
		
	\end{table}

}

	\subsection{Ablation Study}
	In this part, we investigate the performance gain of each proposed module through ablation experiments. All the following experiments are performed on the Cityscapes dataset with $1/8$ labeled data.
	
\begin{figure}[!htbp]
		\begin{center}
			\includegraphics[width=1.0\linewidth]{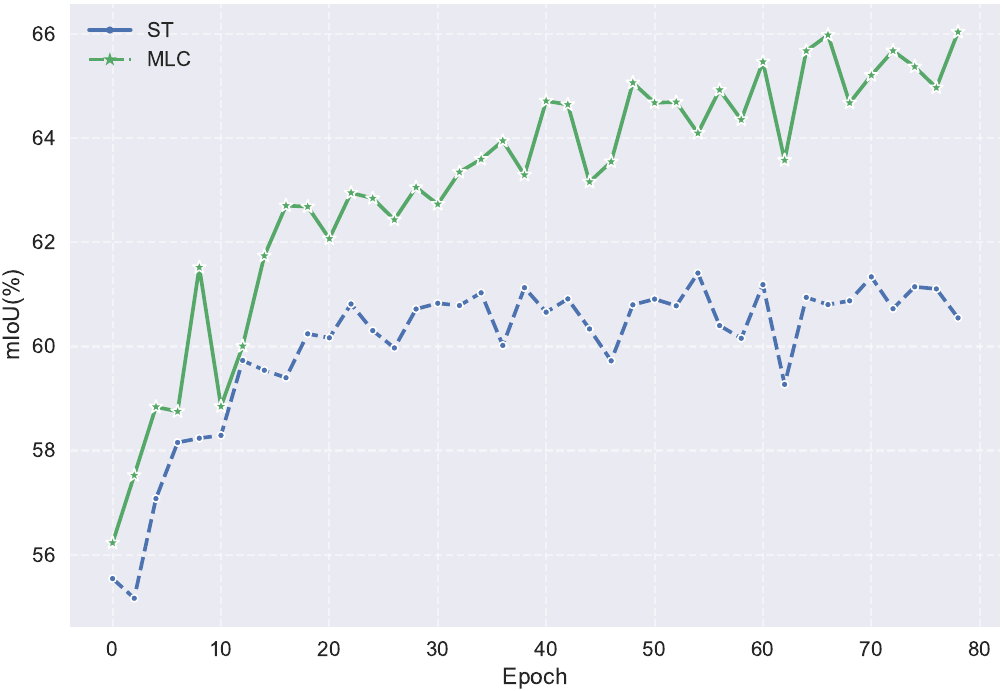}

		\end{center}
		
		\caption{\textbf{Ablation study on multi-level graph.} Quality comparison on pseudo-labels generated by MLLC
			and self-training (ST). 
		}
		\label{trainimage}
		
	\end{figure}

\begin{figure}[!htbp]
		\begin{center}
			\includegraphics[width=1.0\linewidth]{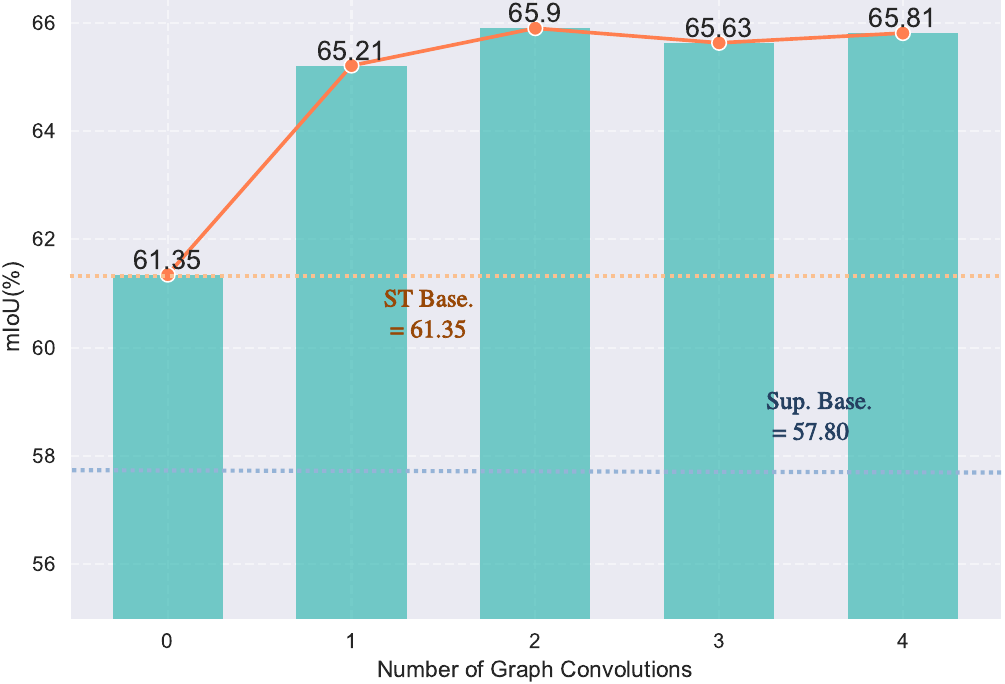}

		\end{center}
		
		\caption{\textbf{Ablation study on multi-level graph.} The effect of different number of graph convolutions $K$. The YELLOW and BLUE horizontal dashed lines indicate the baseline performance of ST and supervised, respectively.
		}
		\label{alaimage}
		
	\end{figure}

	\textbf{Effectiveness of graph-based multi-level label correction.}
	We use ablation experiments to demonstrate the effectiveness of our proposed MLLC. As shown in Fig. \ref{trainimage}(a), our method improves self-training (ST) and produces more accurate pseudo-labels than it. More importantly, as training proceeds MLLC generates more correct pseudo-labels than ST, and the gap between them becomes increasingly large, effectively demonstrating that MLLC can generate more accurate pseudo-labels.
	Fig. \ref{alaimage} shows the importance of graph convolutions on the performance of MLLC. It can be seen that using graph convolutions is effective than without using it. Compared to ST baseline and supervised baseline, MLLC outperforms 4.55\% and 8.10\%, respectively, when using two layer graph convolution. It can also be seen that the performance is not sensitive to the number of convolutions determined by $K$. 
	We also verify performance impact of different update order of SLG and CLG. Intuitively, there are three update orders, i.e., (1) first using the output of the SLG to correct the incorrect pseudolabel of the CLG, and then using the rectified pseudolabel to correct the output of the SLG (CLG first and then SLG), (2) SLG first and then CLG, and (3) SLG and CLG update at the same time. We verify performance effects of different orders in Table \ref{order}. The mIoU of the above three cases are achieved for 65.90\%, 65.16\%, and 64.51\%. Therefore, we select the first case.

	\begin{figure}[!htbp]
		\begin{center}
			
			\includegraphics[width=1\linewidth]{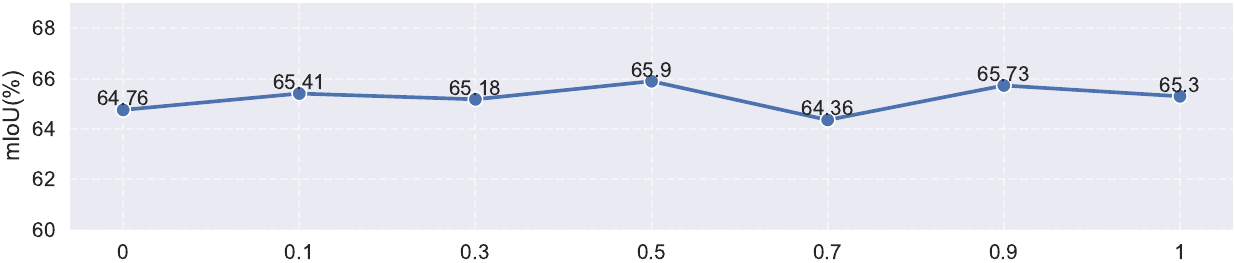}

		\end{center}
		
		\caption{\textbf{Sensitivity analysis for hyper-parameters $\lambda$.}
		}
		\label{senimage}
		
	\end{figure}
	
	\begin{table}[!htbp]

		\caption{Performance impact of the update order of SLG and CLG. $S$, $C$, $\bar{S}$, and $\bar{C}$ denote the original SLG, original CLG, refined SLG, and refined CLG, respectively. The solid line indicates graph convolution, and the dashed line indicates information interaction.
		}
		\centering
		\label{order}
		\resizebox{\linewidth}{!}{
			\begin{tabular}{cccc}  
				
				\toprule
				Order & CLG before SLG & SLG before CLG & SLG \& CLG \\
				\midrule
				Illustrations &  \begin{minipage}{0.2\linewidth} \includegraphics[width=\linewidth]{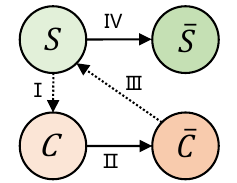} \end{minipage} & \begin{minipage}{0.2\linewidth} \includegraphics[width=\linewidth]{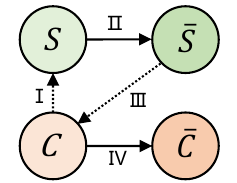} \end{minipage} & \begin{minipage}{0.2\linewidth} \includegraphics[width=\linewidth]{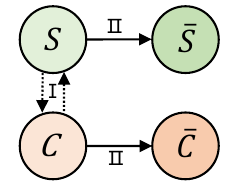} \end{minipage} \\
				\midrule
				mIoU & \textbf{65.90} & 65.16 & 64.51 \\

				\bottomrule
			\end{tabular}
		}
	\end{table}
	
	\textbf{Effectiveness of different components of MLLC.}
	We investigate the effectiveness of each component of the MLLC in Table \ref{abla}. In the table, $Train$ denotes the score of the model on the validation set and $Val.$ denotes the quality of the pseudo-labels generated on unlabeled data. As can be seen, the use of class-specific, intra-image contrastive, inter-image contrastive and dynamic weight can gain 
	performance of 1.01\%, 1.11\%, 0.60\% and 0.49\%, respectively, and when being integrated together,  they can greatly improve the quality of pseudo-labels.
	
	\begin{table}[!htbp]

		\caption{Ablation performance (mIoU \%) of our MLLC method.
		}
		\centering
		\label{abla}
		\resizebox{\linewidth}{!}{
			\begin{tabular}{c|l|cc}  
				
				\toprule
				Methods & Ablations & Train & Val.\\
				\midrule
				Sup. Base. & & - & 57.80 \\
				Self-T. Meth. & & 64.39& 61.35\\
				\midrule
				\multirow{5}{*}{Proposed Meth.} & MLLC w/o class-specific threshold &67.63 &64.89 \\
				& MLLC w/o intra-image contrastive&68.23 & 64.79\\
				&MLLC w/o inter-image contrastive &68.16 & 65.30\\
				&MLLC w/o dynamic weight $\omega_{i,j}^{k}$ & 69.31& 65.41\\
				&MLLC &69.69 &65.90\\
				
				\bottomrule
			\end{tabular}
		}
	\end{table}
	
	\textbf{Impact of different $\text{NN}_k$. 
	}
	We perform comparative experiments using different values of $k$, and the results are shown in Table \ref{ablannk}. When $k$ is zero, no  nearest neighbors are used and it is equivalent to without using the $\text{NN}_k$ approach. As can be seen, when $k>0$, the segmentation performance increases significantly, especially when $k$ is 20. When $k$ is infinity, it corresponds to using a fully connected graphs and the performance is slightly weaker.

	\begin{table}[!htbp]

		\caption{Ablation study on $\text{NN}_k$ in SLG.
		}
		\centering
		\label{ablannk}
		\resizebox{0.85\linewidth}{!}{
			\begin{tabular}{cccccc}  
				
				\toprule
				$\text{NN}_k$ & 0 & 15 & 20 & 25 & $+\infty$\\
				\midrule
				mIoU & 61.35& 65.79 & \textbf{65.90} & 64.93 & 64.36\\

				\bottomrule
			\end{tabular}
		}
	\end{table}
{
	\begin{table}
		\caption{ Comparison between a classic backbone and  a Transformer-based backbone
			.}
		\centering
		\label{trf}
		\begin{tabular}{c|cc}
			\toprule
			Network & Methods & 1/8(372, Cityscapes) \\
			\midrule
			\multirow{2}{*}{DeepLabV3+ \cite{chen2018encoder}} & Supervised Baseline & 71.63 \\
			& MLLC & 77.92 \\
			\midrule
			\multirow{2}{*}{SegFormer \cite{xie2021segformer}} & Supervised Baseline & 74.13 \\
			& MLLC & 79.09 \\
			
			\bottomrule
		\end{tabular}
		
	\end{table}
}
	\textbf{Impact of hyper-parameters $\alpha$.} We examine the segmentation performance of MLLC at different label propagation factors $\alpha$ in Table \ref{ablala}. Intuitively, the ability of the model to correct errors gradually decreases as $\alpha$ increases. The table shows that model is most capable of error correction when $\alpha=1$.
	
	\begin{table}[!htbp]

		\caption{Ablation study on $\alpha$.
		}
		\centering
		\label{ablala}
		\resizebox{\linewidth}{!}{
			\begin{tabular}{ccccccc}  
				
				\toprule
				$\alpha$ & 0 & 0.2 & 0.4 & 0.6 & 0.8 & 1 \\
				\midrule
				mIoU & 65.79 & \textbf{65.90} & 64.83 & 65.71 & 63.71 & 63.14\\

				\bottomrule
			\end{tabular}
		}
	\end{table}
	
	\begin{table}[!htbp]

		\caption{Ablation study on $\sigma$.
		}
		\centering
		\label{ablasig}
		\resizebox{\linewidth}{!}{
			\begin{tabular}{ccccccc}  
				
				\toprule
				$\sigma$ & 0.0 & 0.80 & 0.85 & 0.90 & 0.95 & 1.0 \\
				\midrule
				mIoU & 57.80 & 65.59 & 65.13 & 65.86 & \textbf{65.90} & 65.34\\

				\bottomrule
			\end{tabular}
		}
	\end{table}
	\begin{figure}[!htbp]
		\begin{center}
			\includegraphics[width=1\linewidth]{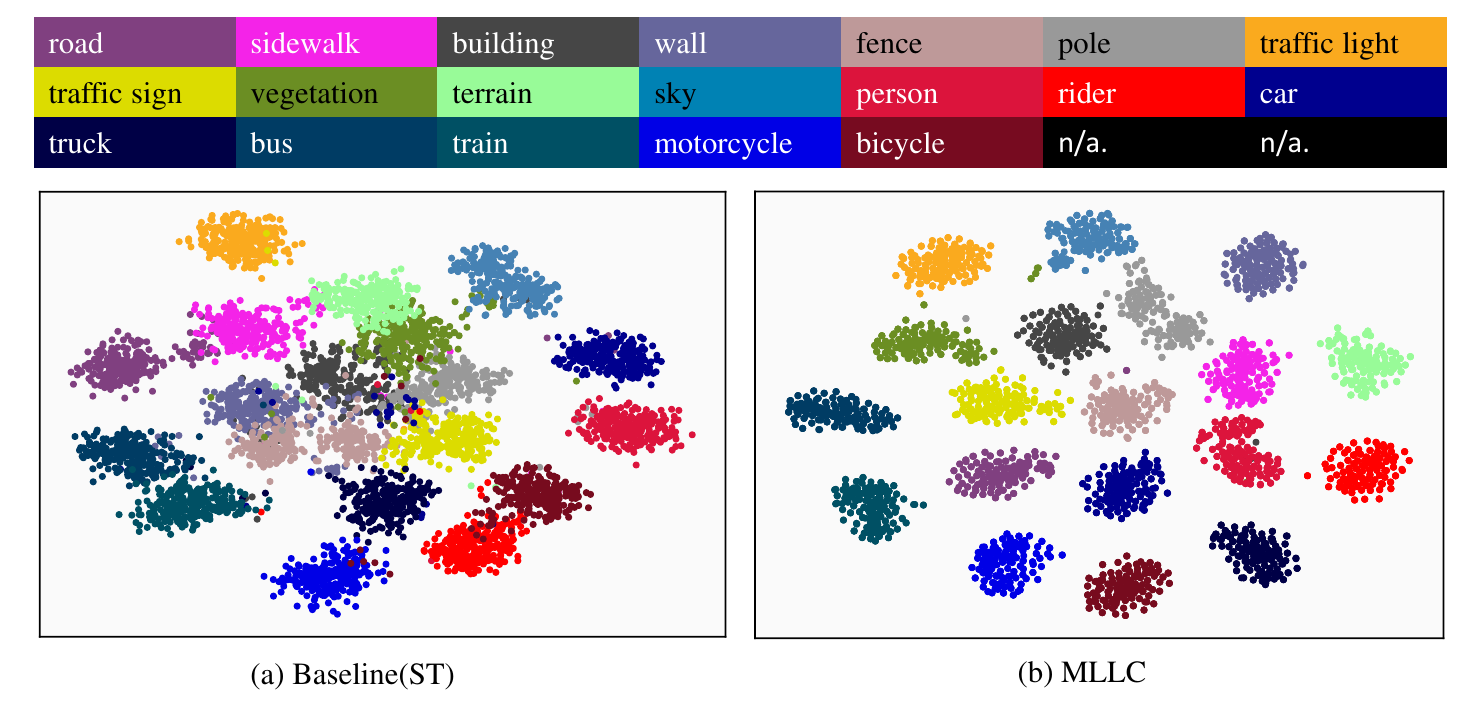}

		\end{center}
		\vspace{-12pt}
		
		\caption{\textbf{Visualization of features, we use t-SNE to map features extracted from input data to a 2D space. We sample 256 points per class for the plot. The results show that MLLC achieves better clustering results.} 
		}
		\label{qualshowf}
		
	\end{figure}
	
	\textbf{Sensitivity analysis of hyper-parameters $\lambda$. } 
	In Fig. \ref{senimage} , we can find that the performance of our approach is not sensitive to a specific value of hyper-parameter $\lambda$ in Eq. (12).

\begin{figure}[!htbp]
		\begin{center}
			\includegraphics[width=1\linewidth]{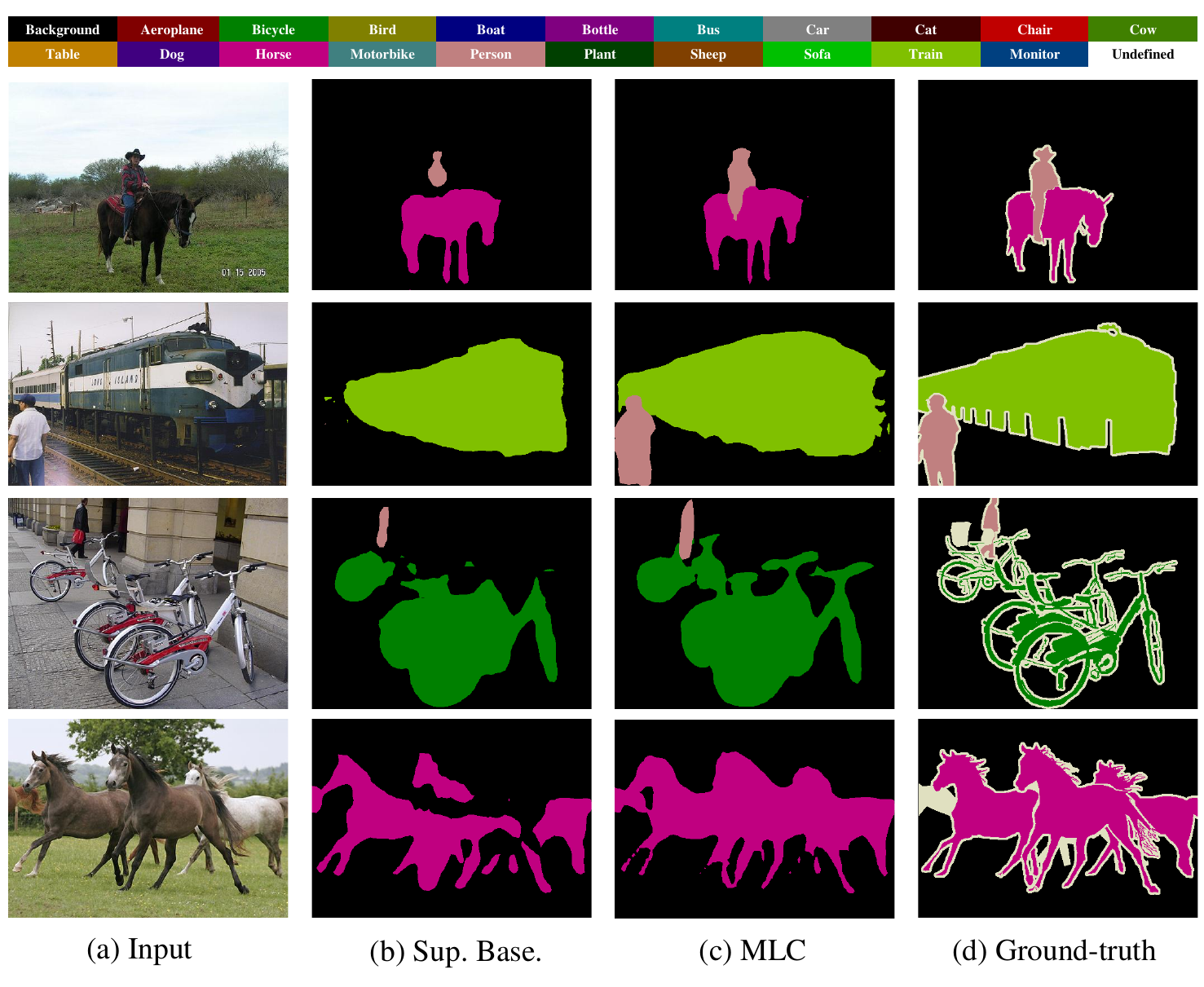}

		\end{center}
		\vspace{-12pt}
		
		\caption{\textbf{Qualitative results on PASCAL VOC 2012 validation set.} 
		}
		\label{qualshow}
		\vspace{-12pt}
		
	\end{figure}

\begin{figure}[!htbp]
		\begin{center}
			\includegraphics[width=1\linewidth]{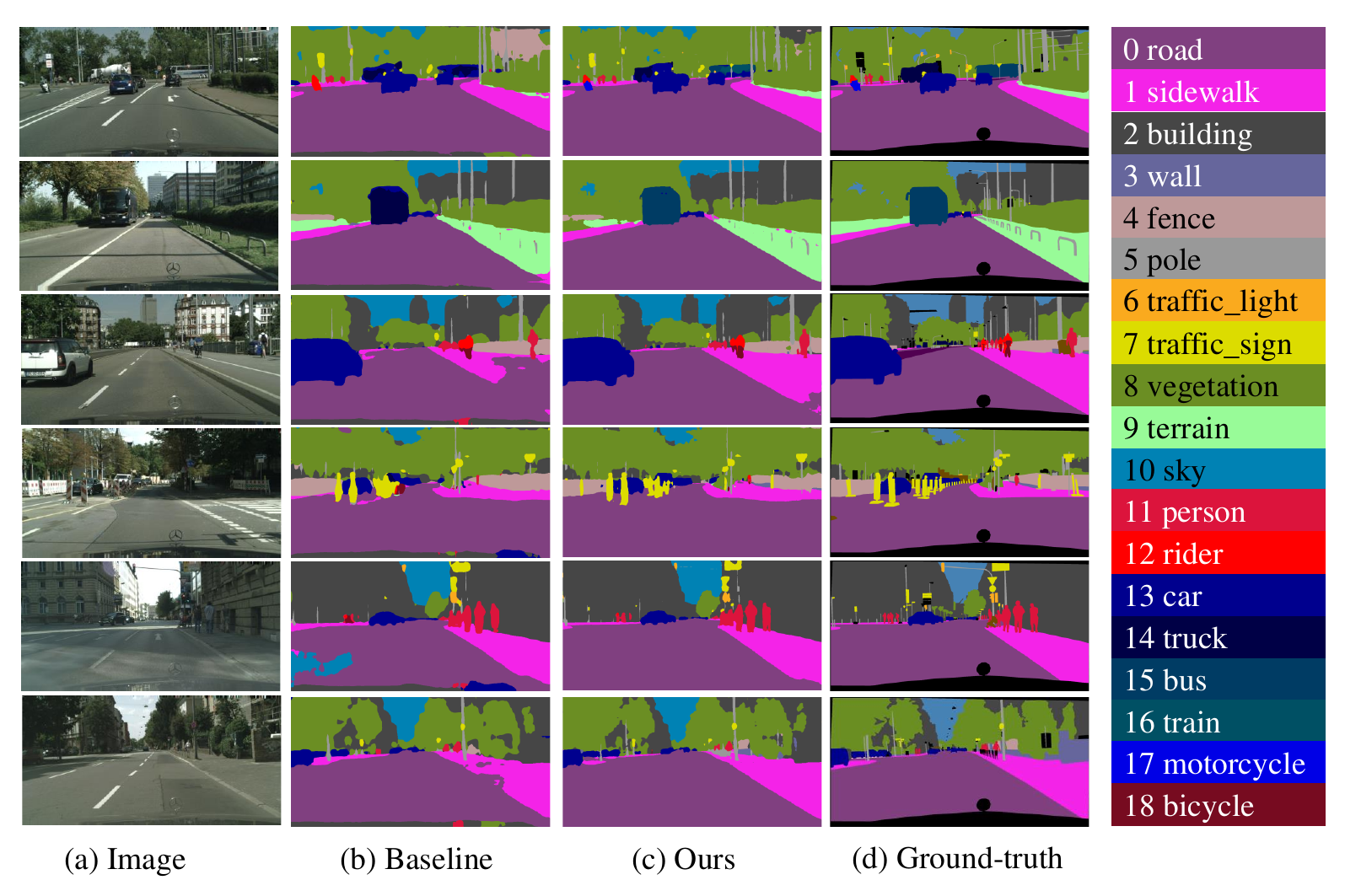}

		\end{center}
		\vspace{-12pt}
		
		\caption{\textbf{Qualitative results on Cityscapes validation set.} 
		}
		\label{qualshowcity}
		\vspace{-12pt}
		
	\end{figure}

	\textbf{Qualitative Results. }
	{ 
	Fig. \ref{qualshowf} shows that MLLC increases the inter-class gaps and reduces the intra-classs distances.
	Fig. \ref{qualshow} and \ref{qualshowcity} show examples of qualitative results on the Pascal VOC 2012 dataset and the Cityscapes dataset, respectively. The segmentation results obtained from the supervised baseline are poor, especially in the person category and small target area. We observe that MLLC has a more complete segmentation compared to supervised baselines and is able to identify some small targets. }

	\section{Conclusion}
	In this paper, we propose a multi-level label correction method by distilling proximate patterns. This approach utilizes graph neural networks to capture structural relationships between semantic-level graphs and class-level graphs to correct incorrect pseudo-labels. It can significantly improve the performance of semi-supervised semantic segmentation, 
	
	and readily achieve new SOTA performance on popular semi-supervised semantic segmentation benchmarks under different partitioning protocols. This framework can be easily used in different backbones as well. 
	
	\section{Acknowledgment}
	This work was supported by Ningbo Science and Technology Innovation Project (No. 2022Z075),  
	the Open Foundation by Zhejiang Key Laboratory of Intelligent Operation and Maintenance Robot (No. SZKF-2022-R03), 
	the Open Fund by Ningbo Institute of Materials Technology \& Engineering, Chinese Academy of Sciences, 
	the National Natural Science Foundation of China (Grant No. 62171244, 61901237), 
	Zhejiang Provincial Natural Science Foundation of China (Grant No. LY23F020011).

	\bibliographystyle{ieeetr}
	\bibliography{ref}

\end{document}